\documentclass[english]{llncs}
\usepackage[T1]{fontenc}
\usepackage[latin9]{inputenc}
\usepackage{amssymb}
\usepackage{stmaryrd}
\usepackage{graphicx}
\usepackage{setspace}
\usepackage[authoryear]{natbib}
\makeatletter

\providecommand{\tabularnewline}{\\}

\makeatother

\usepackage{babel}
\begin{document}
\title{Dataset Mention Extraction in Scientific Articles Using Bi-LSTM-CRF
Model}
\author{Tong Zeng$^{1,2}$ and Daniel Acuna$^{1}$\thanks{Corresponding author: deacuna@syr.edu}}
\institute{$^{1}$School of Information Studies, Syracuse University, Syracuse,
USA\\
$^{2}$School of Information Management, Nanjing University, Nanjing,
China}
\maketitle
\begin{abstract}
Datasets are critical for scientific research, playing an important
role in replication, reproducibility, and efficiency. Researchers
have recently shown that datasets are becoming more important for
science to function properly, even serving as artifacts of study themselves.
However, citing datasets is not a common or standard practice in spite
of recent efforts by data repositories and funding agencies. This
greatly affects our ability to track their usage and importance. A
potential solution to this problem is to automatically extract dataset
mentions from scientific articles. In this work, we propose to achieve
such extraction by using a neural network based on a Bi-LSTM-CRF architecture.
Our method achieves $F_{1}=0.885$ in social science articles released
as part of the Rich Context Dataset. We discuss limitations of the
current datasets and propose modifications to the model to be done
in the future.
\end{abstract}

\section{Introduction}

Science is fundamentally an incremental discipline that depends on
previous scientists' work. Datasets form an integral part of this
process and therefore should be shared and cited as any other scientific
output. This ideal is far from reality: the credit that datasets currently
receive does not correspond to their actual usage \citep{ZENG2020101013}.
One of the issues is that there is no standard for citing datasets,
and even if they are cited, they are not properly tracked by major
scientific indices. Interestingly, while datasets are still used and
mentioned in articles, we lack methods to extract such mentions and
properly reconstruct dataset citations. The Rich Context Competition
challenge aims to close this gap by inviting scientists to produce
automated dataset mention and linkage detection algorithms. In this
chapter, we detail our proposal to solve the dataset mention step.
Our approach attempts to provide a first approximation to better give
credit and keep track of datasets and their usage.

The problem of dataset extraction has been explored before. Ghavimi
et al. \citeyearpar{ghavimiIdentifyingImprovingDataset2016,ghavimiSemiautomaticApproachDetecting2017}
use a relatively simple TF-IDF representation with cosine similarity
for matching dataset identification in social science articles. Their
method consists of four major steps: preparing a curated dictionary
of typical mention phrases, detecting dataset references, and ranking
matching datasets based on cosine similarity of TF-IDF representations.
This approach achieved a relatively high performance, with $F_{1}=0.84$
for mention detection and $F_{1}=0.83$, for matching. \citet{singhalDataExtractMining2013}
proposed a method using normalized Google distance to screen whether
a term is in a dataset. However, this method relies on external services
and is not computationally efficient. They achieve a good $F_{1}=0.85$
using Google search and $F_{1}=0.75$ using Bing. A somewhat similar
project was proposed by \citet{luDatasetSearchEngine2012}. They built
a dataset search engine by solving the two challenges: identification
of the dataset and association to a URL. They build a dataset of 1000
documents with their URLs, containing 8922 words or abbreviations
representing datasets. They also build a web-based interface. This
shows the importance of dataset mention extraction and how several
groups have tried to tackle the problem.

In this article, we describe a method for extracting dataset mentions
based on a deep recurrent neural network. In particular, we used a
Bidirectional Long short-term Memory (Bi-LSTM) sequence to sequence
model paired with a Conditional Random Field (CRF) inference mechanism.
We tested our model on a novel dataset produced for the Rich Context
Competition challenge. We achieve a relatively good performance of
$F_{1}=0.885$. We discuss the noise and duplication present in the
dataset and limitations of our model.

\section{The dataset}

The Rich Context Dataset challenge was proposed by the New York University's
Coleridge Initiative \citep{richtextcompetition}. The challenge comprised
several phases, and participants moved through the phases depending
on their performance. We only analyse data from the first phase. This
phase contained a list of datasets and a labelled corpus of around
5000 publications. Each publication was labelled indicating whether
a dataset was mentioned within it and which part of the text mentioned
it. The challenge used an accuracy measure for measuring the performance
of the competitors and also the quality of the code, documentation,
and efficiency.

We adopted the CoNLL 2003 format \citep{tjong2003introduction} to
annotate whether a token is a part of dataset mention. Concretely,
we use the tag DS to denote a dataset mention; a B- prefix indicates
that the token is the beginning of a dataset mention, an I- prefix
indicates that the token is inside a dataset mention, and O denotes
a token that is not a part of a dataset mention. We put each token
and its tag (separated by a horizontal tab control character) on one
line, and use the end-of-line (\textbackslash n) control character
as separator between sentences (see Table \ref{tab:tagging-format}).
The dataset was randomly split by 70\%, 15\%, 15\% for training set,
validation set and testing set, respectively.

\begin{table}

\caption{\protect\label{tab:tagging-format}Example of a sentence annotated
by IOB tagging format}

\begin{centering}
\begin{tabular}{ll}
\hline 
Token & Annotation\tabularnewline
\hline 
This & O\tabularnewline
\ldots{} & \tabularnewline
data & O\tabularnewline
from & O\tabularnewline
the & O\tabularnewline
Monitoring & B-DS\tabularnewline
the & I-DS\tabularnewline
Future & I-DS\tabularnewline
( & O\tabularnewline
MTF & B-DS\tabularnewline
) & O\tabularnewline
\textbackslash n & \tabularnewline
\hline 
\end{tabular}
\par\end{centering}
\end{table}

\section{The Proposed Method}

\subsection{Overall view of the architecture}

In this section, we propose a model for detecting mentions based on
a Bi-LSTM-CRF architecture. At a high level, the model uses a sequence-to-sequence
recurrent neural network that produces the probability of whether
a token belongs to a dataset mention. The CRF layer takes those probabilities
and estimates the most likely sequence based on constrains between
label transitions (i.e., mention--to--no-mention--to-mention has
low probability). While this is a popular architecture for modeling
sequence labeling, the application to our particular dataset and problem
is new. 

\begin{figure}
\begin{centering}
\includegraphics[width=1\textwidth]{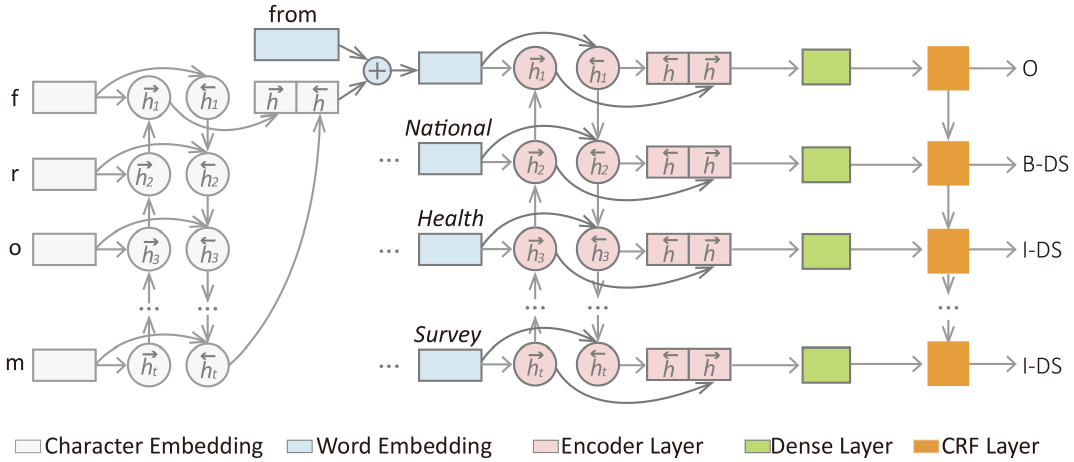}
\par\end{centering}
\caption{\protect\label{fig:NetworkArchitecture}Network Architecture of Bi-LSTM-CRF
network}
\end{figure}

We now describe in more detail the choices of word representation,
hyper-parameters, and training parameters. A schematic view of the
model is in Fig \ref{fig:NetworkArchitecture} and the components
are as follows:
\begin{enumerate}
\item Character embedding layer: treat a token as a sequence of characters
and encode the characters by using a bidirectional LSTM to get a vector
representation.
\item Word embedding layer: mapping each token into fixed sized vector representation
by using a pre-trained word vector.
\item One Bi-LSTM layer: make use of Bidirectional LSTM network to capture
the high level representation of the whole token sequence input.
\item Dense layer: project the output of the previous layer to a low dimensional
vector representation of the the distribution of labels.
\item CRF layer: find the most likely sequence of labels.
\end{enumerate}

\subsection{Character Embedding}

Similar to the bag of words assumption, a word could be composed of
characters sampled from a bag of characters. Previous research \citep{pmlr-v32-santos14,jozefowicz2016exploring}
has shown that the use of character-level embedding could benefit
multiple NLP-related tasks. In order to use character-level information,
we break down a word into a sequence of characters, then build a vocabulary
of characters. We initialize the character embedding weights using
the vocabulary size of a pre-defined embedding dimension, then update
the weights during the training process to get the fixed-size character
embedding. Next, we feed a sequence of the character embedding into
an encoder (a Bi-LSTM network) to produce a vector representation
of a word. By using a character encoder, we can solve the out-of-vocabulary
problem for pre-trained word embedding, as every word could be composed
of characters.

\subsection{Word Embedding}

The word embedding layer is responsible for storing and retrieving
the vector representation of words. Specifically, the word embedding
layer contains a word embedding matrix $M^{tkn}\in\mathbb{R}^{d|V|}$,
where $V$ is the vocabulary of the tokens and $d$ is the dimension
of the embedding vector. The embedding matrix was initialized by a
pre-trained GloVe vectors \citep{pennington2014glove}, and updated
by learning from the data. In order to retrieve from the embedding
matrix, we first convert a given sentence into a sequence of tokens,
then for each token we look up the embedding matrix to get its vector
representation. Finally, we get a sequence of vectors as input for
the encoder layer.

\subsection{LSTM}

A recurrent neural network (RNN) is a type of artificial neural network
which recurrently takes the output of the previous step as input of
the current step. This recurrent nature allows it to learn from sequential
data, for example, the text which consists of a sequence of works.
An RNN could in theory capture contextual information in variable-length
sequences, but it suffers from gradient exploding/vanishing problems
\citep{pascanu2013difficulty}. The long short-term memory (LSTM)
architecture was proposed by \citet{hochreiter1997long} to cope with
these gradient problems. Similar to a standard RNN, the LSTM network
also has a repeating module called an LSTM cell. The cell remembers
information over arbitrary time-steps because it allows information
to flow along it without change. The cell state is regulated by a
forget gate and an input gate which control the proportion of information
to forget from a previous time-step and to remember for a next time-step.
Also, there is an output gate controlling the information flowing
out of the cell. The LSTM could be defined formally by the following
equations:

\begin{equation}
i_{t}=\sigma(W_{i}x_{t}+W_{i}h_{t-1}+b_{i})
\end{equation}
\begin{equation}
f_{t}=\sigma(W_{f}x_{t}+W_{f}h_{t-1}+b_{f})
\end{equation}
\begin{equation}
g_{t}=tanh(W_{g}x_{t}+W_{g}h_{t-1}+b_{g})
\end{equation}
\begin{equation}
o_{t}=\sigma(W_{o}x_{t}+W_{o}h_{t-1}+b_{o})
\end{equation}
\begin{equation}
c_{t}=f_{t}\varotimes c_{t-1}+i_{t}\varotimes g_{t}
\end{equation}
\begin{equation}
h_{t}=o_{t}\varotimes tanh(c_{t})
\end{equation}
where $x_{t}$ is the input at time $t$, $W$ is the weights, $b$
is the bias. The $\sigma$ is the sigmoid function, $\varotimes$
denotes the dot product, $c_{t}$ is the LSTM cell state at time $t$
and $h_{t}$ is hidden state at time $t$. The $i_{t}$, $f_{t}$,
$o_{t}$ and $g_{t}$ are named as input, forget, output and cell
gates respectively, they control the information to keep in its state
and pass to next step.

LSTM can learn from the previous steps, which is the left context
if we feed the sequence from left to right. However, the information
in the right context is also important for some tasks. The bi-LSTM
\citep{graves2013speech} satisfies this information need by using
two LSTMs. Specifically, one LSTM layer was fed by a forward sequence
and the other by a backward sequence. The final hidden states of each
LSTM were concatenated to model the left and right contexts:

\begin{equation}
h_{t}=\left[\overrightarrow{h_{t}}\varoplus\overleftarrow{h_{t}}\right]
\end{equation}

Finally, the outcomes of the states are taken by a CRF layer \citep{lafferty2001conditional}
that takes into account the transition nature of the beginning, intermediate,
and end of mentions.

\section{Results}

In this work we wanted to propose a model for the Rich Context Competition
challenge. We propose a relatively standard architecture based on
the bi-LSTM CRF network. We now describe the evaluation metrics, hyperparameter
setting, and the results of this network on the dataset provided by
the competition. 

For all of our results, we use F1 as the measure of performance. This
measure is the harmonic average of the precision and recall and it
is the standard measure used in sequence labelling tasks. It varies
from 0 to 1, the higher the better. Our method achieved a relatively
high $F_{1}$ of 0.885 for detecting mentions.

\begin{table}

\caption{\protect\label{tab:hyperparameter}Model search space and best assignments}

\centering{}%
\begin{tabular}{lll}
\hline 
Hyperparameter & Search space & Best parameter\tabularnewline
\hline 
Number of epochs & 50 & 50\tabularnewline
Patience & 10 & 10\tabularnewline
Batch size & 64 & 64\tabularnewline
Pre-trained word vector size & choice{[}50, 100, 200,300{]} & 100\tabularnewline
Encoder hidden size & 300 & 300\tabularnewline
Number of encoder layers & 2 & 2\tabularnewline
Dropout rate & choice{[}0.0,0.5{]} & 0.5\tabularnewline
Learning rate optimizer & Adam & Adam\tabularnewline
L2 regularizer & 0.01 & 0.01\tabularnewline
Learning rate & 0.001 & 0.001\tabularnewline
\hline 
\end{tabular}
\end{table}

We train models using the training data and monitor the performance
using the validation data (we stop training if the performance does
not improve for the last 10 epochs). We use the Adam optimizer with
learning rate 0.001 and batch size equal to 64. The hidden size of
LSTM for character and word embedding is 80 and 300, respectively.
For the regularization methods, and to avoid overfitting, we use L2
regularization set to 0.01 and we also use dropout rate equal to 0.5.
We trained eight models with a combination of different GloVe vector
size (50, 100, 300 and 300) and dropout rate (0.0, 0.5). The hyperparameter
settings are shown in Table \ref{tab:hyperparameter}.

The test performances are reported in Table 11.3. The best model is
trained by word vector size 100 and dropout rate 0.5, with F1 score
0.885 (Table \ref{tab:Performance-of-proposed}), and it takes 15
hours 58 minutes for the training on an NVIDIA GTX 1080 Ti GPU in
a computer with an Intel Xeon E5-1650v4 3.6 GHz CPU with 128GB of
RAM.

\begin{table}
\caption{\protect\label{tab:Performance-of-proposed}Performance of proposed
network}

\centering{}%
\begin{tabular}{cccccc}
\hline 
Models & GloVe size & Dropout rate & Precision & Recall & $F_{1}$\tabularnewline
\hline 
m1 & 50 & 0.0 & 0.884 & 0.873 & 0.878\tabularnewline
m2 & 50 & 0.5 & 0.877 & 0.888 & 0.882\tabularnewline
m3 & 100 & 0.0 & 0.882 & 0.871 & 0.876\tabularnewline
m4 & 100 & 0.5 & 0.885 & 0.885 & \textbf{0.885}\tabularnewline
m5 & 200 & 0.0 & 0.882 & 0.884 & 0.883\tabularnewline
m6 & 200 & 0.5 & 0.885 & 0.880 & 0.882\tabularnewline
m7 & 300 & 0.0 & 0.868 & 0.886 & 0.877\tabularnewline
m8 & 300 & 0.5 & 0.876 & 0.878 & 0.877\tabularnewline
\hline 
\end{tabular}
\end{table}

We also found some limitations to the dataset. Firstly, we found that
mentions are nested (e.g. HRS, RAND HRS, RAND HRS DATA are linked
to the same dataset). The second issue most of the mentions have ambiguous
relationships to datasets. In particular, only 17,267 (16.99\%) mentions
are linked to one dataset, 15,292 (15.04\%) mentions are listed to
two datasets, and 12,624 (12.42\%) are linked to three datasets. If
these difficulties are not overcome, then the predictions from the
linkage process will be noisy and therefore impossible to tell apart.

\section{Conclusion}

In this work, we report a high-accuracy model for the problem of detecting
dataset mentions. Because our method is based on a standard Bi-LSTM-CRF
architecture, we expect that updating our model with recent developments
in neural networks would only benefit our results. We also provide
some evidence of how difficult we believe the linkage step of the
challenge could be if dataset noise is not lowered.

One of the shortcomings of our approach is that the architecture is
lacking some modern features of RNN networks. In particular, recent
work has shown that attention mechanisms are important especially
when the task requires spatially distant information, as in this case.
These benefits could also translate to better linkage. We are exploring
new architectures using self-attention and multiple-head attention.
We hope to share these approaches in the near future.

There are a number of improvements that we could make in the future.
A first improvement would be to use non-recurrent neural architectures
such as the Transformer which has been shown to be faster and a more
effective learner than RNNs. Another improvement would be to bootstrap
information from other dataset sources such as open-access full-text
articles from PubMed Open Access Subset. This dataset contains dataset
citations \citep{ZENG2020101013} -- in contrast to the most common
types of citations to publications. The location of such citations
within the full text could be exploited to perform entity recognition.
While this would be a somewhat different problem than the one solved
in this article, it would still be useful for the goal of tracking
dataset usage. In sum, by improving the learning techniques and the
dataset size and quality, we could significantly increase the success
of finding datasets in publications.

Our proposal, however, is surprisingly effective. Because we have
barely modified a general RNN architecture, we expect that our results
will generalize relatively well either to the second phase of the
challenge or even to other disciplines. We would emphasize, however,
that the quality of the dataset has a great deal of room for improvement.
Given how important this task is for the whole of science, we should
strive to improve the quality of these datasets so that techniques
like this one can be more broadly applied.

\section*{Acknowledgements}

Tong Zeng was funded by the China Scholarship Council \#201706190067.
Daniel E. Acuna was funded by the National Science Foundation awards
\#1646763 and \#1800956.

\bibliographystyle{apalike}
\bibliography{rcc-06}

\begin{thebibliography}{}

\bibitem[{Coleridge Initiative}, 2019]{richtextcompetition}
{Coleridge Initiative} (2019).
\newblock Rich context competition.

\bibitem[Ghavimi et~al., 2017]{ghavimiSemiautomaticApproachDetecting2017}
Ghavimi, B., Mayr, P., Lange, C., Vahdati, S., and Auer, S. (2017).
\newblock A semi-automatic approach for detecting dataset references in social
  science texts.
\newblock {\em Information Services \& Use}, 36(3-4):171--187.

\bibitem[Ghavimi et~al., 2016]{ghavimiIdentifyingImprovingDataset2016}
Ghavimi, B., Mayr, P., Vahdati, S., and Lange, C. (2016).
\newblock Identifying and {{Improving Dataset References}} in {{Social Sciences
  Full Texts}}.
\newblock {\em arXiv:1603.01774 [cs]}.

\bibitem[Graves et~al., 2013]{graves2013speech}
Graves, A., Mohamed, A.-r., and Hinton, G. (2013).
\newblock Speech recognition with deep recurrent neural networks.
\newblock In {\em Acoustics, speech and signal processing (icassp), 2013 ieee
  international conference on}, pages 6645--6649. IEEE.

\bibitem[Hochreiter and Schmidhuber, 1997]{hochreiter1997long}
Hochreiter, S. and Schmidhuber, J. (1997).
\newblock Long short-term memory.
\newblock {\em Neural computation}, 9(8):1735--1780.

\bibitem[Jozefowicz et~al., 2016]{jozefowicz2016exploring}
Jozefowicz, R., Vinyals, O., Schuster, M., Shazeer, N., and Wu, Y. (2016).
\newblock Exploring the limits of language modeling.
\newblock {\em arXiv preprint arXiv:1602.02410}.

\bibitem[Lafferty et~al., 2001]{lafferty2001conditional}
Lafferty, J., McCallum, A., and Pereira, F.~C. (2001).
\newblock Conditional random fields: Probabilistic models for segmenting and
  labeling sequence data.

\bibitem[Lu et~al., 2012]{luDatasetSearchEngine2012}
Lu, M., Bangalore, S., Cormode, G., Hadjieleftheriou, M., and Srivastava, D.
  (2012).
\newblock A {{Dataset Search Engine}} for the {{Research Document Corpus}}.
\newblock In {\em 2012 {{IEEE}} 28th {{International Conference}} on {{Data
  Engineering}}}, pages 1237--1240, Arlington, VA, USA. {IEEE}.

\bibitem[Pascanu et~al., 2013]{pascanu2013difficulty}
Pascanu, R., Mikolov, T., and Bengio, Y. (2013).
\newblock On the difficulty of training recurrent neural networks.
\newblock In {\em International Conference on Machine Learning}, pages
  1310--1318.

\bibitem[Pennington et~al., 2014]{pennington2014glove}
Pennington, J., Socher, R., and Manning, C. (2014).
\newblock Glove: Global vectors for word representation.
\newblock In {\em Proceedings of the 2014 conference on empirical methods in
  natural language processing (EMNLP)}, pages 1532--1543.

\bibitem[Santos and Zadrozny, 2014]{pmlr-v32-santos14}
Santos, C.~D. and Zadrozny, B. (2014).
\newblock Learning character-level representations for part-of-speech tagging.
\newblock In Xing, E.~P. and Jebara, T., editors, {\em Proceedings of the 31st
  International Conference on Machine Learning}, volume~32 of {\em Proceedings
  of Machine Learning Research}, pages 1818--1826, Bejing, China. PMLR.

\bibitem[Singhal and Srivastava, 2013]{singhalDataExtractMining2013}
Singhal, A. and Srivastava, J. (2013).
\newblock Data {{Extract}}: {{Mining Context}} from the {{Web}} for {{Dataset
  Extraction}}.
\newblock {\em International Journal of Machine Learning and Computing}, pages
  219--223.

\bibitem[Tjong Kim~Sang and De~Meulder, 2003]{tjong2003introduction}
Tjong Kim~Sang, E.~F. and De~Meulder, F. (2003).
\newblock Introduction to the conll-2003 shared task: Language-independent
  named entity recognition.
\newblock In {\em Proceedings of the seventh conference on Natural language
  learning at HLT-NAACL 2003-Volume 4}, pages 142--147. Association for
  Computational Linguistics.

\bibitem[Zeng et~al., 2020]{ZENG2020101013}
Zeng, T., Wu, L., Bratt, S., and Acuna, D.~E. (2020).
\newblock Assigning credit to scientific datasets using article citation
  networks.
\newblock {\em Journal of Informetrics}, 14(2):101013.

\end{thebibliography}

\end{document}